*Commentary*

# PDDL 2.1: Representation vs. Computation

**Héctor Geffner** HECTOR.GEFFNER@UPF.EDU
*ICREA – Universitat Pompeu Fabra*
*Paseo de Circunvalacion 8*
*08003 Barcelona, Spain*

## Abstract

I comment on the PDDL 2.1 language and its use in the planning competition, focusing on the choices made for accommodating time and concurrency. I also discuss some methodological issues that have to do with the move toward more expressive planning languages and the balance needed in planning research between semantics and computation.

## 1. Introduction

Fox and Long should be thanked and congratulated for their effort in organizing and running the 3rd International Planning Competition. They came up with an extended planning language along with a number of new problems and domains that challenged existing planners and will certainly challenge future planners as well. The goal of this note is to comment on some of the decisions that went into the design of the language (Fox & Long, 2003). I'll organize these comments into three parts: the extensions and their motivation, the choices made for accommodating time and concurrency, and the uses of the language in the competition.

## 2. The Extensions and their Motivation

The Planning Domain Description Language, PDDL 1, was developed to encourage "sharing problems and algorithms" and "meaningful comparison of performance", and more specifically to "provide a (common) notation for problems in the AIPS-98 Planning Contest" (McDermott, 2000). PDDL 1 borrowed features from a number of existing planners, yet only a subset of the language was used in the competition: mostly the Strips and ADL subsets along with types. Features like axioms, safety constraints, and expression evaluations were not used, and actually didn't make it into the PDDL document of the second competition (Bacchus, 2001). For the third competition, significant extensions were introduced including

1. Numeric fluents (in addition to boolean fluents)
2. Actions with durations
3. Arbitrary plan metrics
4. Continuous change

The introduction of these features appeared to be guided by two main criteria: whether the *semantics* of the extensions could be stated clearly and whether they would enable the encoding of "realistic problems". Computational considerations seemed to have played a lesser role. Indeed, while at the time of the first competition, Strips and ADL planning were well understood and several computational approaches had been developed, the third competition featured extensions that aimed to present new functionalities that few if any reported planners could handle. More precisely, while computational approaches for dealing with numeric fluents and actions with durations have been studied for some time (e.g., Koehler, 1998; Rintanen & Jungholt, 1999; Hoffmann, 2002; Laborie & Ghallab, 1995; Smith & Weld, 1999; Jonsson, Morris, Muscettola, & Rajan, 2000; Haslum & Geffner, 2001), proposals for dealing with arbitrary plan metrics and continuous change have been more concerned with semantics than with computation (e.g., McDermott, 2003).

Setting new challenges in the form of more general types of planning tasks is necessary and positive for the field, although there are costs as well. AI planning, like AI itself, is a big 'elephant', and progress for a small community is bound to become slow and scattered if one researcher looks at a leg, a second at the trunk, and a third at the tail. Focus is important, and focus on the basics is even more important (see the progress in SAT solving for example; e.g. Berre & Simon, 2003). A continuous shift in the 'problem to be solved', while showing a genuine concern for more powerful models and applications may also deter the accumulation of knowledge that is necessary for achieving solid progress. The transition of SAT and CSP ideas from the lab to the real world took many years but they would have probably taken many more, if these communities hadn't kept a focus on the basic problems to be solved.

Likewise, challenges come in different sizes and some are more basic than others. Moreover, some may be well-defined semantically but not computationally. E.g., do we expect useful and general computational methods that will work for *any plan metric*? Most likely not: an approach suitable for reaching the goal with a minimum number of actions, may not be good for reaching the goal with minimum resource consumption or with a maximum number of goods. These may be completely different computational problems even if their description is very similar (e.g., just think about replacing an addition sign by a multiplication sign in the cost function of a linear program). If a planner can tackle such generic problems, we should probably be suspect about the quality of the solutions it finds, as mostly likely, it is moving from one solution to the next *blindly* (i.e., with no estimation of plan costs). Similarly, while cost structures given by the sum of action costs provide a simple and approachable generalization of classical planning, an analogous additive structure in which action costs depend not only on the action but also on the *state* on which the action is done, is considerably more difficult. Namely, the difficulty in finding plans remains unchanged, but the difficulty in finding good or optimal plans increases, as without good pruning rules or estimators, one is bound to consider one complete plan after another.

### 2.1 Time, Resources, and Concurrency

The addition of numeric fluents and time in PDDL 2.1 is a sensible move that aims to incorporate features existent in a number of planners (e.g., Laborie & Ghallab, 1995; Smith & Weld, 1999; Jonsson et al., 2000). While it's not possible to express arbitrary constraints between actions and/or variables as in some of these planners, the basic functionality and computational challenges are there. An important omission in my view is the absence of explicit *resources,* separate from action pre and postconditions. Resources in scheduling and some planning systems (e.g., Wilkins, 1988; Currie & Tate, 1991; Baptiste, Pape, & Nuijten, 2001) are used to determine the level of *concurrency* that is allowed in a problem. E.g., if 7 workers are available, and there are 4 tasks that require 2 workers each, then at most 3 of these tasks can be executed concurrently. Otherwise, the resources needed exceed the capacity. This is an example of a multi-capacity, renewable resource; other types of resources are common (e.g., unary resources, consumable resources, etc). In PDDL 2.1, resources need to be encoded as numeric fluents, and the syntax of the operators through some agreed upon conventions (more about this below), determines the level of concurrency allowed. In my view, this choice is unfortunate for three reasons:

- *heuristically*, because explicit resources can be exploited computationally in a way that general numeric fluents can't,
- *semantically*, because explicit resources provide an account of concurrency that is simple and clean, something that cannot be said about the form of concurrency in PDDL 2.1 or Graphplan, and
- *conceptually*, because explicit resources, along with time, provide the natural generalization and unification of planning and scheduling.

Let us start with the first issue. Resources are used to control the level of concurrency among tasks: if explicitly accommodated in the language, they don't have to appear in action pre or postconditions or in the goal. For resources encoded as fluents, the opposite is true. The result is that, unless resources are automatically identified by domain analysis, pruning mechanisms and lower bounds developed for handling resources (e.g., Laborie, 2003) can't be used.

Likewise, the definition of concurrency in terms of resources is transparent. A set of actions $A$ can be executed concurrently at time $t$ if the resources needed by these actions do not exceed the capacity available at $t$. On the other hand, in PDDL 2.1, as in Graphplan, the level of concurrency is defined implicitly in terms of the *syntax* of pre and postconditions. In Graphplan, for example, two operators can be executed concurrently if they don't interfere with each other; i.e., if one does not delete preconditions or positive effects of the other. PDDL 2.1 extends this definition in a number of ways, taking into account the duration of actions and the intervals over which preconditions have to be preserved. For simplicity, I'll focus on Graphplan's notion of concurrency only, leaving these extensions aside.

Consider the Blocks World and the actions $move(a, b, c)$ and $move(a, b, d)$ that move a block $a$ from $b$ to $c$ and $d$ respectively. Clearly, these actions cannot be done concurrently as a block cannot be moved to two different destinations at the same time. They are indeed mutex in Graphplan (and PDDL 2.1) as both have a precondition ($on(a, b)$) that they delete. Yet, why should deleting a precondition of an action $a$ prevent an action $a'$ from being executed concurrently if the action $a$ itself deletes the precondition? The justification for this notion of concurrency has never been made explicit. In Graphplan it was adopted because it ensures that all serializations of a set of concurrent, applicable actions remain applicable and yield the same result (Blum & Furst, 1995). Yet, why should this same criterion be the most convenient in a truly concurrent setting?

Consider now the actions $move(a, b, c)$ and $move(d, e, f)$ which do not interfere, and thus are deemed concurrent. This is a correct assumption to make if the number of robot arms is sufficiently large, but is incorrect otherwise. Of course the encoding can be fixed by playing with the operators; e.g., in the presence of a single arm, the *stack/unstack* encoding would provide the correct assumption of concurrency, while in the presence of three arms, a similar encoding, more involved, would be possible as well. In any case, the account of concurrency based on action interference as defined in Graphplan and PDDL 2.1 carries certain implicit assumptions and the question is whether we want to make those assumptions, and whether they are reasonable or not.

The account of concurrency based on resources is more transparent in this sense, and makes heuristic information for computing feasible plans more explicit. The account also provides a good degree of flexibility,[1] although in certain contexts cannot replace the need for more general constraints (e.g., to say that robot arms cannot collide or cannot get too close).

## 3. PDDL and the Planning Competition

The authors of PDDL 2.1 are motivated by the goal of "closing the gap between planning research and applications". Since they cannot influence applications, they presumably mean that they want PDDL 2.1 to have an influence on planning research, moving it away from toy domains toward realistic applications. No one can object to a move toward realistic applications, yet, I think, it's useful to keep in mind that 'toy domains' have served the AI Problem Solving community very well: Sliding puzzles in Heuristic Search, the n-Queens in Constraint Satisfaction, Blocks World in Planning, etc. They have provided focus, and conceptually simple problems that are computationally challenging as well. This mix is convenient for identifying simple but powerful ideas: heuristic estimators, pattern databases, constraint propagation and global constraints, and so on. I believe it is worth asking, from this perspective, whether a toy problem like Blocks World is exhausted. It clearly makes for bad press, but have we learned to solve the problem well, in a domain-independent fashion? If not, we may be still missing some fundamental ideas that are likely to be needed in richer settings also featuring complex interactions between actions pre and postconditions. Such problems are most often intractable, yet if the performance gap between general planners and specialized solvers remains large, the applicability of planning technology will remain limited. Thus while it is positive and necessary to make room in PDDL and the Planning Competition for more complex planning languages and tasks, I believe that it is also necessary to identify and maintain a focus on some core computational problems.

The competition and its supporting language should also give researchers and practitioners an idea of the state of the field: what kind of problems can be modeled and solved, what kind and size of problems can be solved well, what approaches work best and when, and also how fast or slow the field is progressing. Progress means better solutions in less time, and if in addition, a commitment to optimality is made, progress means more powerful

1. E.g., in the Blocks World, the blocks themselves can be regarded as renewable, unary resources which are needed by actions like moving, painting, and cleaning blocks. The result is that all these actions will be mutex. Similarly, the arm can be treated as a resource, and then all actions requiring the arm will be regarded as mutex, etc.

ideas for pruning larger parts of the search space safely and effectively.[2] The competitions held so far have contributed to these goals, and in this way, have contributed to the progress of the field and to its empirical grounding. Still, care should be taken so that these lessons can be drawn from future competitions as well. A basic requirement in the presence of powerful modeling languages such as PDDL 2.1, is to separate the basic functionalities into different tracks so that planners are not rewarded only by their coverage, but also by how well they do in each class of tasks. At the same time it would be useful to distinguish tracks that reflect mature research from tracks that feature research at an early stage. And paraphrasing the 'no-moving-target rule' in the lead article, I think that we should maintain the focus on the 'basic problems to be solved' and resist the AI urge to move on to new problems as soon as the original problems begin to crack.

2. Safely here means that at least one optimal solution is *not* pruned; this guarantee is not needed when the optimality requirement is dropped.